\documentclass[sigconf]{aamas} 

\usepackage{balance} 
\usepackage{amsfonts}
\usepackage{multirow}
\usepackage{romannum}
\usepackage[ruled,vlined]{algorithm2e}
\usepackage{hyperref}

\setcopyright{ifaamas}
\copyrightyear{2021}
\acmYear{2021}
\acmDOI{}
\acmPrice{}
\acmISBN{}

\acmSubmissionID{???}

\title[AAMAS-2021 Formatting Instructions]{Selective Intervention Planning using Restless Multi-Armed Bandits to Improve Maternal and Child Health Outcomes}

\author{Siddharth Nishtala$^{1}$, Lovish Madaan$^{2}$, Aditya Mate$^{3\dagger}$, Harshavardhan Kamarthi$^{1}$, Anirudh Grama$^{2}$,} 
\author{Divy Thakkar$^{2}$, Dhyanesh Narayanan$^{2}$, Suresh Chaudhary$^{4}$, Neha Madhiwalla$^{4}$, Ramesh Padmanabhan$^{4}$, Aparna Hegde$^{4}$, Pradeep Varakantham$^{2}$, Balaraman Ravindran$^{1}$, Milind Tambe$^{2,3}$}
\affiliation{
  \institution{$\;$\\$^{1}$Robert Bosch Centre for Data Science and Artificial Intelligence, IIT Madras\\
  $^{2}$Google Research\\
  $^{3}$Harvard University\\
  $^{4}$ARMMAN}
}

\begin{abstract}
India has a maternal mortality ratio of 113 and child mortality ratio of 2830 per 100,000 live births. Lack of access to preventive care information is a major contributing factor for these deaths, especially in low resource households. We partner with ARMMAN, a non-profit based in India employing a call-based information program to disseminate health-related information to pregnant women and women with recent child deliveries. We analyze call records of over 300,000 women registered in the program created by ARMMAN and try to identify women who might not engage with these call programs that are proven to result in positive health outcomes. We built machine learning based models to predict the long term engagement pattern from call logs and beneficiaries' demographic information, and discuss the applicability of this method in the real world through a pilot validation. Through a pilot service quality improvement study, we show that using our model's predictions to make interventions boosts engagement metrics by 61.37\%. We then formulate the intervention planning problem as restless multi-armed bandits (RMABs), and present preliminary results using this approach.\footnotetext{$\dagger$Work done during an internship at Google Research}
\end{abstract}

\newcommand{\BibTeX}{\rm B\kern-.05em{\sc i\kern-.025em b}\kern-.08em\TeX}

\begin{document}


\pagestyle{fancy}
\fancyhead{}


\maketitle
\section{Introduction}

Two of the eight Millennium Development Goals (MDGs), which were adopted by all United Nations Member States in 2000, were to reduce child mortality and to improve maternal health. Despite significant progress being made on both fronts, the improvements made did not meet the set targets for 2015. The achieved under-five mortality rate of 43 per 1,000 live births fell short of the target of 30 per 1,000 live births, and the achieved maternal mortality ratio of 210 per 100,000 births was well below the set target of 95 per 100,000 births \cite{mdg-ritchie}. It was observed that almost 86\% of the estimated maternal deaths in 2013 were accounted for by developing countries in sub-Saharan Africa and Southern Asia \cite{mdg-un}. Most of these deaths are preventable and can be managed by providing care and support in the antenatal and postnatal periods and skilled care during childbirth. 

As of 2014, only 52\% of all pregnant women in developing regions received the recommended number of antenatal visits. This creates a problem of information asymmetry where pregnant women are not equipped with preventive care information. Many organizations have begun using information and communications technology (ICT) to bridge these gaps and mitigate health risks of both the mother and her child. ARMMAN is one such organization that runs a free voice-call based program named mMitra that delivers preventive care information in a timely manner to educate and empower women with critical information during pregnancy and after delivery to help the enrolled women identify and mitigate pregnancy related risks. The program consists of 141 pre-recorded calls sent over a period of 21 months in the language and time of choice of the enrolled woman.

However, about 40\% of the enrolled women fail to listen to more than 30 seconds in half of the calls that they answer, leaving both the woman and her child at risk. Providing interventions in the form of calls from health workers could potentially help retain a larger fraction of beneficiaries in the program. Existing methods use a variety of machine learning models to find beneficiaries that are likely to drop out of the program \cite{nishtala2020missed}. While these methods provide a set of beneficiaries to intervene on, the resource limitations in the field often make it difficult to intervene on all beneficiaries flagged by the models. Thus, there is immense value in building pipelines that not only identify beneficiaries that are likely to drop out of the program, but also learn from data generated from interventions to identify beneficiaries that are most likely to benefit from interventions.

Our contributions are three fold in this paper. First, we consider the problem of forecasting the beneficiaries' long-term engagement based on their demographic information and past engagement with the voice call program. We formulate this task as a supervised learning problem and build machine learning based models that are trained on previous call-related data of over 300,000 beneficiaries to accurately predict the long-term engagement. 

Second, we conduct a Pilot Service Quality Improvement Study (PSQIS) on a subset of beneficiaries who are predicted to be low engaging for validating our models from the prediction task. 

\begin{figure}[h]
    \centering
    \includegraphics[width=\linewidth]{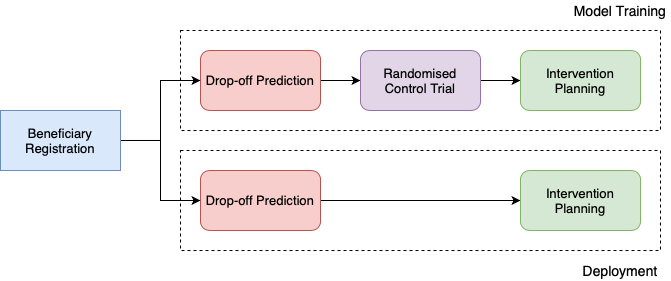}
    \caption{Training and Deployment Pipelines for ARMMAN's mMitra program}
    \label{fig:pipelines}
\end{figure}

Finally, using the intervention data from the PSQIS experiment, we develop a Restless Multi-Arm Bandit (RMAB) based framework for planning interventions over multiple decision epochs with limited resources. We do this to maximize the effects of the interventions by selecting a subset of the predicted low engagement beneficiaries who are likely to change their behaviour and engage more with the program, if an intervention is done on them. The role of each contribution in the training and deployment pipelines is detailed in Figure \ref{fig:pipelines}.

\section{Related Work}


Patient adherence is a very important problem in the context of healthcare. For some categories of diseases, more than 40\% of patients sustain long-term health risks by ignoring and misunderstanding health advice \cite{martin2005challenge}. This can lead to increased healthcare costs and increased burden on medical professionals, hospitals, and the government.

Adherence in patients has been studied for a variety of diseases like heart problems \cite{son2010application}, HIV \cite{HIV}, and Tuberculosis \cite{10.1001/archinte.1996.00440020063008, Killian_2019}. \cite{son2010application} uses Support Vector Machines (SVMs) to identify predictors or features that convey adherence information for heart failure patients. \cite{Killian_2019} builds machine learning models to predict adherence in Tuberculosis patients and use this information to provide interventions to the patients so that they adhere to medicinal doses. These studies use actual medicinal adherence data for the analysis of patient health outcomes, whereas we operate in the call-based healthcare domain where automated calls providing important information on maternal and child health are placed to the beneficiaries registered in the program. Adherence or engagement in call-based health programs is difficult to sustain because of issues like patient interest, call connection failure, etc.

Previous studies like \cite{nishtala2020missed} use demographic and call features to build CNN and LSTM models to predict beneficiaries who are at high risk of drop-off from the call-based program. NGO and healtcare workers can then provide message or call interventions to these predicted high-risk beneficiaries and encourage them to engage with the program and improve the overall health outcomes. However, NGOs like ARMMAN usually have limited resources and are not able to call all the predicted high-risk beneficiaries. So a key challenge arising out of the prediction of beneficiaries having higher probability of low engagement, is selecting a subset of these low engaging beneficiaries to intervene on. This intervention planning problem has applications in many areas such as device maintenance \cite{villar-machine}, anti-poaching patrols \cite{QianRestlessPoachers, xu2020dualmandate}, and healthcare \cite{MateCollapsingBandits, mate2021risk}.

We formulate the intervention planning problem as a Restless Multi-Armed Bandit (RMAB), and the subset of beneficiaries for intervention are selected according to their Whittle Indices \cite{whittle-rbs}. Some recent works \cite{LiuIndexRB, MateCollapsingBandits, QianRestlessPoachers, mate2021risk} have shown the applicability of RMABs for intervention planning in other domains while providing asymptotic performance guarantees for the solutions obtained.
\section{Data Description}
We use the registration information of women beneficiaries registered in the \textit{mMitra} program along with their automated call records. The registration information includes demographic data such as their age, education levels, income groups, phone owner in the family, etc. The registration information also includes the date they were registered in the program, their gestation age at the time of enrollment, preferred language and call slots for the automated / intervention calls. The automated call records contain a mapping from the beneficiary user IDs to the corresponding call records, where each call contains the message ID, call date, duration and a flag corresponding to whether the call was successful or not. We use the message ID as the identifier for the gestation age of the beneficiary since messages for beneficiaries with different gestation ages are different.

\textbf{Call Statistics:} All the registration and call data is anonymized and contains call records for 329,489 beneficiaries registered in 2018. The total number of calls are 70,272,621. It might be the case that the calls are not able to go through to the beneficiaries, either due to connection failure or the beneficiary not picking up the call. Due to this, ARMMAN attempts two calls for each connection failure. We consider the call with the maximum duration for each beneficiary for our analysis and this results in a total of 27,488,890 call records.

\textbf{Terminology:} Each call attempt to a beneficiary is called an \textbf{Attempt}. A successful \textit{attempt} resulting in the beneficiary picking up the call is called a \textbf{Connection}. If the beneficiary stays on the call for more than 30 seconds, it is called an \textbf{Engagement}.

We also deal with the ratio of number of engagements to the number of connections in further sections for the definition of risk. This ratio tells us about the engagement of beneficiaries with the program. In later sections, we refer to this simply as the E2C Ratio.

\section{Long-term Engagement Prediction}

We build machine learning models that predict if a woman would not engage with the program in the long-term given the call history of one month from their registration in the program and the demographic information. These predictions can help identify beneficiaries who require more involvement from health care workers, and plan for long-term interventions such as providing access to counselors.

\subsection{Problem Formulation}
For each beneficiary registered in the program and having atleast 8 months of call records, we use the demographic information and one month of call records as the input features. The remaining call records are used to calculate the output label for the problem. Since predicting long-term engagement requires us to consider a longer period of time for generating the output labels, we generate labels that disentangle the effect of connection failure by considering the E2C Ratio. After calculating the ratio, we define a risk threshold that helps us binarize the ratio into high long-term engagement (HLTE) and low long-term engagement (LLTE). For example, if a beneficiary has engaged 5 out of the 15 times that the connection was successful, the engagements to connections (E2C) ratio is 0.33 and on binarizing this with a engagement threshold of 0.5, the beneficiary is labelled as \texttt{LLTE}.


Choosing one month of features allows us to make predictions fairly early in the program without compromising much on the performance. We additionally require that a beneficiary has a minimum of 24 connections in the prediction period. Our final dataset contains data for 143,356 beneficiaries.

It is important to note that the value of engagement threshold dictates the amount of imbalance in the classes. We use a threshold of 0.5 which results in 97,029 HLTE beneficiaries and 46,327 LLTE beneficiaries. We also learned that a model with high recall for low engaging beneficiaries would be more beneficial for ARMMAN. Therefore, to account for the imbalance in the dataset and to prioritize high recall on the low engaging beneficiaries, we train models using different class weights for the HLTE and LLTE classes.



\subsection{Models}
For the baseline, we provide a rule-based model: beneficiaries having E2C Ratio less than 0.5 in the first one month are assigned as LLTE and HLTE otherwise. The rule-based model basically predicts the beneficiary will have similar behaviour in the future as it is in the first month after her registration.

We also use a random forest model over all demographic and call features. The random forest model consists of 200 trees with a maximum depth 30. For the neural baselines, we use two neural networks called CoNDiP (Convolutional Neural Disengagement Predictor) and ReNDiP (Recurrent Neural Disengagement Predictor) from \cite{nishtala2020missed}.

CoNDiP has 1D Convolutional \cite{kiranyaz20191d} layers to encode the call features. The call features are also called dynamic features because of the variable number of calls to each beneficiary. We keep an upper limit of 8 call attempts to the beneficiaries in one month and append zeros to call sequences having less than 8 call attempts. We use 2 convolutional layer with 8 kernels of size 3. We also do average pooling after each layer. We use a single transformation matrix for the static features. Outputs from the final layer of the CNN and the transformed static features are concatenated and another transformation matrix is used to generate the final output label. The transformation matrices defined above are single-layered MLPs.

In ReNDiP, the call data is encoded using recurrent layers because of their sequential nature. The CNN sub-architecture in CoNDiP is replaced with a bi-directional LSTM \cite{hochreiter1997long} for ReNDiP. There are 8 units in the LSTM layer and the hidden layer size is 100. Outputs from the LSTM and transformed static features are concatenated and passed to another MLP of size 10 and then finally the output label is generated.

Both CoNDiP and ReNDiP were trained with class weights of 1 and 1.75 for HLTE and LLTE classes respectively.

\begin{table}[h]
\centering
\caption{Results for long-term engagement task on 2018 registrations}
\begin{tabular}{|c|c|c|c|c|}
\hline
Model  & Accuracy & Precision & Recall & F1 score \\ \midrule
RF$^{\mathrm{a}}$     & 0.791     & 0.710      & 0.601   & 0.651     \\ \hline
CoNDiP & 0.798     & 0.680      & 0.711   & 0.695     \\ \hline
ReNDiP & 0.791     & 0.656      & 0.744   & 0.697     \\ \hline
Rule-Based & 0.754 & 0.604 & 0.703 & 0.649 \\ \hline
\multicolumn{5}{l}{$^{\mathrm{a}}$RF: Random Forest}
\end{tabular}
\label{tab:long-term-engagemnt}
\end{table}




\subsection{Results}
The results for the engagement prediction task are summarized in Table \ref{tab:long-term-engagemnt}. We see that CoNDiP, ReNDiP and RF perform better than the baseline in terms of accuracy and recall. Based on ARMMAN's requirement of using the models with a higher recall for the LLTE class, we use the ReNDiP model for the deployment pipeline.

\section{Deployment Study}

\subsection{Phase 1: Model Validation}
The health workers from ARMMAN have access to an easy-to-use dashboard to visualize the beneficiaries' predicted long-term engagement. We piloted our ReNDiP prediction model and the dashboard app on beneficiaries registered in November 2019.

Using ReNDiP to study the relevance of the predictions, we gathered the available call data for these beneficiaries, removed the call logs of the first 30 days (as they were used to build input features) and calculated the E2C ratio for each beneficiary. To correctly analyse the performance of the model, we filter out beneficiaries with zero connections. The results are summarized in Table \ref{tab:pilot-overall}.

\begin{table}[h]
\centering
\caption{Results for long-term engagement task on November 2019 registrations}
\begin{tabular}{|c|c|c|c|c|}
\hline
Model$^{\mathrm{a}}$ & Accuracy & Precision & Recall & F1 score \\ \midrule
ReNDiP  & 0.730     & 0.667      & 0.708\footnotemark   & 0.687     \\ \hline
\end{tabular}
\label{tab:pilot-overall}
\end{table}
\footnotetext[1]{We are investigating an ensemble of approaches from Table \ref{tab:long-term-engagemnt} to improve the recall even further}


\subsection{Phase 2: PSQIS}

The ReNDiP model's predictions are used to make interventions on beneficiaries predicted as LLTE, for ensuring the beneficiaries' continued engagement with the program. To assess the utility of the model and the interventions that follow, we conduct a Pilot Service Quality Improvement Study (PSQIS) with four groups. First is the control group where no interventions are made on the beneficiaries. The second group consists of beneficiaries that had a text message sent to their registered phone numbers through Short Message Service (SMS). The third group includes beneficiaries that were called by \textit{ARMMAN}'s call center executives. For the fourth and final group, we use a strategy for interventions that is a combination of SMS and call interventions. For beneficiaries in this group, we begin by sending an SMS on the first day of interventions, and after observing their behavior for six weeks, we make a call intervention for beneficiaries that did not exhibit a significant change in behavior after the SMS intervention. The motivation behind this strategy was to avoid call interventions (which require a higher amount of time and effort) for beneficiaries that responded well to SMS interventions.

We consider beneficiaries that were registered in May, June and July 2020. This included a total of 25,169 beneficiaries that satisfied the constraints set on the number of attempts. We then use the ReNDiP model from Section 4 to make predictions and identify LLTE beneficiaries. From \textit{ARMMAN}'s past observations from the field, it was found that calls made to phone numbers that hardly registered any past engagement, were often to a wrong person. To channel the available resources towards beneficiaries that are more likely to benefit from the interventions, we filter away beneficiaries that had less than two engagements in the first sixty days. This brought down the total number of beneficiaries in the intervention pool to 6563 beneficiaries. These beneficiaries were distributed across the four arms while ensuring that the demographic features and behavior across the groups largely remained the same. The number of beneficiaries in control, SMS, call and SMS+call groups were 1663, 1597, 1627 and 1676 respectively. The interventions were started on 23 October 2020 and the behavior of the beneficiaries was observed until 14 February 2021.


To evaluate the effects of the interventions, we consider a new evaluation metric of post-intervention engagement behavior. Since we only have the beneficiaries' call data to make any observations, we again use the E2C Ratio for this metric. A beneficiary is said to exhibit high engagement behavior if her E2C ratio computed over all calls in the post-intervention period is greater than 0.5. This metric is also computed over a shorter period of 15 weeks as opposed to the long-term engagement (LTE) labels defined in Section 4, which are computed over a period of 7 months. In Table \ref{tab:conversions-groups-p1}, we show the percentages of predicted LLTE beneficiaries that exhibit high engagement behavior in the post-intervention period. We observe a significant increase in the percentage of beneficiaries exhibiting high engagement behavior in the call group.

\begin{table}[htbp]
\centering
\caption{Percentages of predicted LLTE beneficiaries that exhibit high engagement behavior in the post-intervention period.}
\begin{tabular}{|c|c|c|c|c|c|}
\hline
 & Control & SMS & Hybrid & Call \\ \hline
\% High Eng.   & 23.3\%  & 27.7\% & 29.7\% & 37.6\% \\ \hline
\end{tabular}
\label{tab:conversions-groups-p1}
\end{table}

 We observe that SMS interventions are not as effective compared to other modes in encouraging beneficiaries to engage with the program. In our analysis of the interventions, we observed that the success rate of call interventions is 45.2\%. Despite a seemingly low success rate, we observe that call interventions are very effective. Comparing the call intervention group to the control group, we see a relative improvement of 61.37\% in percentage of high engagement beneficiaries in the post-intervention period. The pilot study suggests that the AI model made reliable predictions on beneficiaries that are likely to drop out, and coupling the model's predictions with interventions decreases the proportion of beneficiaries that exhibit low engagement in the program.
 
 We also note that call interventions planned for beneficiaries may not always be successful, i.e beneficiaries might not receive calls due to network failures. To understand the impact of successful interventions, we also calculate the results for beneficiaries that were successfully intervened. 43.6\% of beneficiaries in the successful call group exhibit high post-intervention engagement behaviour. However, the success or failure of the intervention may also be affected by factors that may influence the behavior of the beneficiaries in both pre-intervention and post-intervention periods. Hence, we find it necessary to report results both for the entire intervention group, and the subset of beneficiaries that got a intervention successfully.

\section{Intervention Planning}
Beneficiaries registered in the program do not have a linear trend in engagement, and drop-off and re-engage with varying patterns. While the prediction model gives us beneficiaries that are likely to have low engagement in the long term, it does not help us identify a subset of beneficiaries that are likely to benefit the most from an intervention at any given time. To derive maximum benefit from the available resources, there is immense value in accounting for re-engagement patterns and learning from interventions made in the past while scheduling interventions for low engaging beneficiaries in the future.

To facilitate continuous interventions on low-engagement beneficiaries with resource limitations (in conducting interventions), we represent this problem as an Restless Multi-Arm Bandit (RMAB). In the RMAB, the planner has to pick $k$ out of $N$ arms at each time step for intervention, where each arm evolves according to a Markov Decision Process (MDP), and the goal is to maximize the cumulative reward over all time steps.
In order to transfer knowledge on impact of intervention across beneficiaries, we cluster beneficiaries when computing transition probabilities. All beneficiaries in the same cluster have the same MDP parameters for transitions.


The RMAB system forms the final component in the pipeline that flags beneficiaries for interventions. The pool of $N$ beneficiaries considered are the ones flagged as LLTE by the long-term engagement prediction model. The motivation behind this was to allocate our budget among the beneficiaries known to be "low engagement" listeners in the program. This reduces the computational costs of following the behavior of all beneficiaries throughout the duration of the program. 

\subsection{Restless Multi-Armed Bandits}
An RMAB consists of $N$ independent 2-action Markov Decision Processes (MDP) \cite{mdp-puterman}, where each MDP is defined by the tuple $\{\mathcal{S}, \mathcal{A}, r, \mathcal{P}\}$. $\mathcal{S}$ denotes the state space, $\mathcal{A}$ is the set of possible actions, $r$ is the reward function $r : \mathcal{S} \times \mathcal{A} \times \mathcal{S} \rightarrow \mathbb{R}$ and $\mathcal{P}$ represents the transition function, with $P_{s, s'}^{a}$ as the probability of transitioning from state $s$ to state $s'$ under the action $a$. $\pi : \mathcal{S} \rightarrow \mathcal{A}$ represents the MDP policy and consists of a mapping from the state space to the action space specifying which action to take in a particular state. In our formulation, we use the discounted reward criterion for measuring the return of a policy $\pi$. Starting from an initial state $s_0$, the discounted reward is defined as $R_{\beta}^{\pi}(s_0) =\sum_{t = 0}^{\infty}\beta^tr(s_t)$, where $\beta \in [0,1)$ is the discount factor and the actions are selected according to the policy $\pi$. The state value function $V^{\pi} (s_t)$ is then defined as the expected return starting from the state $s_t$ and following the policy $\pi$. Similarly, the action value function $Q^{\pi} (s_t, a_t)$ is defined as the expected return starting from $s_t$, taking the action $a_t$ and following the policy $\pi$.

Each independent MDP in the RMAB formulation is controlled by an arm and the total return of the planning setup is the sum of the total individual returns of each of the arms. The goal here is to maximize the total expected reward summed up across all arms.

Computing the optimal policy for an RMAB is PSPACE-hard. To deal with this, Whittle \cite{whittle-rbs} proposed an index-based heuristic, known today as the Whittle Index, that can be solved in polynomial time. It has been shown to be asymptotically optimal for the time average reward problem \cite{weber1990index} and other families of RMABs arising from stochastic scheduling problems \cite{glazebrook2006indexable}.

The Whittle Index technique operates by computing an index for every arm, which intuitively captures the value of pulling that arm. The Whittle approach considers each arm separately for the index computation, effectively decoupling the arms and converting it to $N$ smaller problems. The Whittle Index policy for the RMAB is to pull the $k$ arms with the highest index values.

The Whittle Index approach hinges around the key idea of a  ``passive subsidy'', which is a hypothetical reward offered to the planner for choosing the passive action, in addition to the original reward function. We use $Q_m^{\pi} (s_t, a_t)$ to denote the expected return starting from $s_t$, taking the action $a_t$ and following the policy $\pi$, when the passive subsidy offered is $m$. Intuitively,  higher the passive subsidy (denoted as $m$) offered, more attractive is the passive action to the planner.  Whittle Index is defined as the infimum value of this subsidy that makes the planner indifferent between the `active' and the `passive' actions. Formally, the Whittle Index is defined as $W(s)=inf_m\{m:Q_m(s,a=0)=Q_m(s,a=1)\}$

Such a passive subsidy value that makes both the actions equally attractive to the planner is well-defined only if a technical condition called `indexability' is satisfied. This condition also guarantees the asymptotic optimality of the Whittle Index heuristic. In practice, Whittle index has proven to be a useful heuristic in many settings even when indexability has not been formally proven. We use a similar approach in this paper, leaving a proof of indexability for future work.

\subsection{Intervention Planning as RMAB}
Formally, we represent the problem of identifying $k$ beneficiaries to call and intervene at each time step out of $N$ beneficiaries to maximize the amount of collective cumulative engagement across the program’s entire duration as a Restless Multi-Armed Bandit (RMAB) problem. 

\begin{figure}[h]
    \centering
    \includegraphics[width=\linewidth]{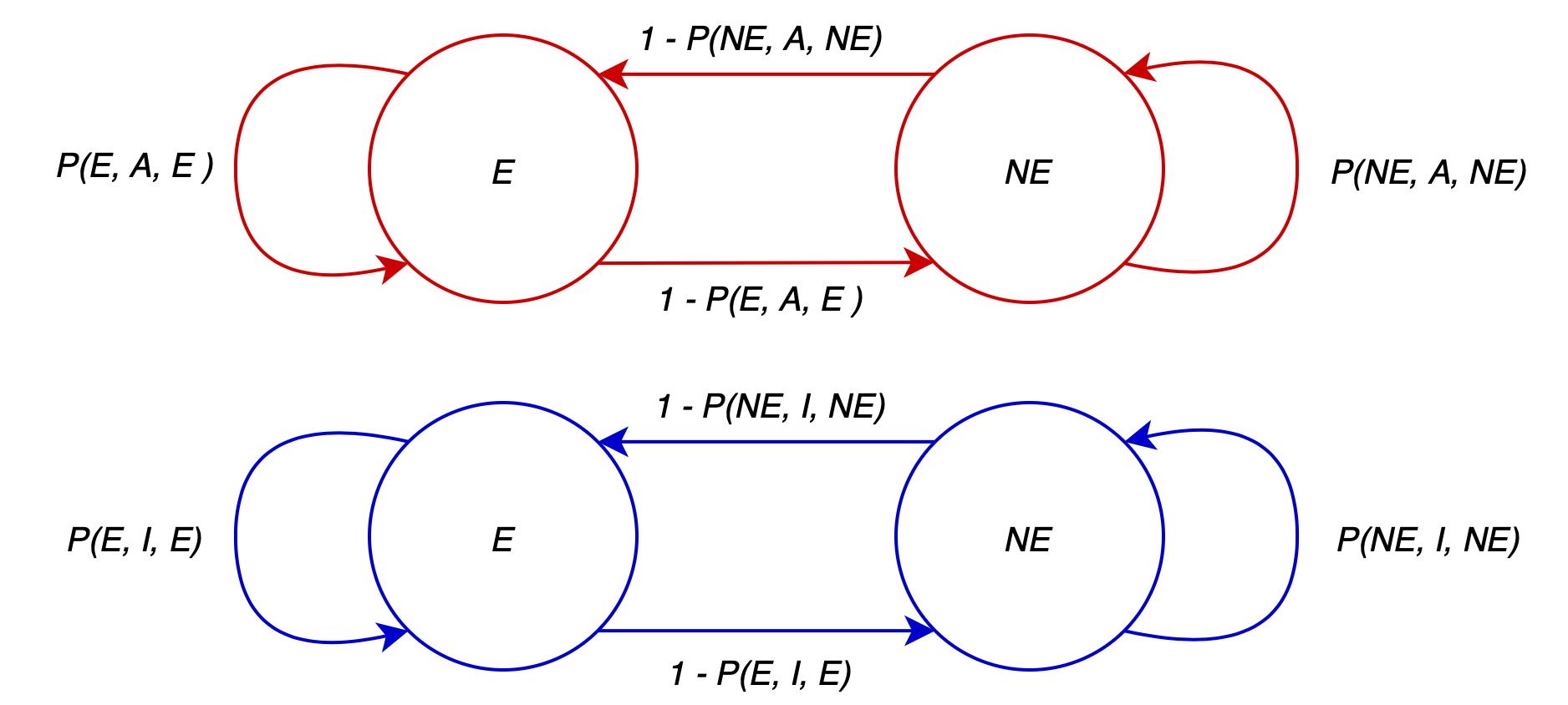}
    \caption{MDP for Restless Bandits. The upper graph (in red) corresponds to action $A$ and the lower graph (in blue) corresponds to action $I$.}
    \label{fig:R-MDP}
\end{figure}

The engagement behavior of each beneficiary is dictated by an underlying MDP. Every beneficiary is considered an arm, and pulling an arm corresponds to the action of attempting a call intervention. We name this action as $I$. The passive action of abstaining from intervening is named $A$. The state is then defined to account for the recent engagement behavior of the beneficiary. We do this by calculating the E2C ratio over calls in the previous one month and using a 0.5 threshold to assign one of two states: either engaging ($E \equiv E2C \geq 0.5$) or not engaging ($NE \equiv E2C < 0.5$). Figure \ref{fig:R-MDP} describes the MDP for each beneficiary. The beneficiary behavior is characterized by the transition probabilities: $P(E, A, E)$, $P(NE, A, NE)$, $P(E, I, E)$, and $P(NE, I, NE)$. We refer to these transition probabilities also as the parameters of the MDP, that are learnt from the data. The reward function of the MDP is defined on the basis of the current state the beneficiary: reward of +1 if the beneficiary is in the $E$ state, and -1 if the beneficiary is in the $NE$ state. Hence, high engagement over a period of time would correspond to a high positive return and low engagement would correspond to a high negative return.

\subsection{Methods}

We use the data generated from the PSQIS experiment to learn a policy. We partition the available data into two sets: (1) a training set for estimating the parameters of the MDPs and learning a policy through the RMABs framework, and (2) a test set to retrospectively evaluate the performance of the learned system.

\subsubsection{Parameter Estimation of the MDPs}
We begin by following each beneficiary through the program, and computing the state after each month by considering the E2C ratio over the calls made in that month. This gives us a sequence of states that summarize the behavior of the beneficiary over the course of the program. We then note any interventions that may have been made and include this information as actions in the state sequence. This gives us a sequence of the form: $s_1, a_1, s_2, a_2$ and so on, where $s_i \in \{E, NE\}$ and $a_i \in \{I, A\}$. We construct this sequence for every beneficiary in the training set. We then break these sequences down into 3-tuples of $(s_t, a_t, s_{t+1})$. Each tuple corresponds to a transition made by a beneficiary.

As noted earlier, grouping beneficiaries together allows us to capture the effects corresponding to both actions on the behavior of the beneficiaries. Therefore, we group beneficiaries with comparable demographic and call features in the first month to estimate the parameters of the MDPs. All transitions corresponding to the beneficiaries in a group are pooled and used to estimate the parameters of the MDP corresponding to that group. However, estimating parameters for each group in this way could have a drawback. A few groups have fewer beneficiaries than others, and hence have fewer transitions to estimate parameters from. This could result in noisy parameters. To overcome this issue, we use K-means clustering \cite{lloyd1982least} to cluster groups with similar parameters together and recompute a pooled estimate by considering the transitions of all beneficiaries in the cluster. The number of clusters is a hyper-parameter that can be tuned to find the best behavior clusters within beneficiaries. Note that all beneficiaries belonging to a cluster share the parameters corresponding to the cluster.

\subsubsection{Computation of Whittle Index}
We use the estimated parameters to compute the value function for each MDP through value iteration \cite{Sutton1998}. Given that all beneficiaries in a cluster share the MDP parameters of the cluster they belong to, it suffices to compute the value function for each cluster. This reduces computational costs significantly as the cost does not scale with respect to the number of beneficiaries in consideration but with respect to the number of clusters. Since the MDP has no terminal states, we introduce a discount factor to ensure the convergence of value iteration. The Whittle Index can then be calculated as the value of $m$ that satisfies: $Q_m(s_t, a_t=I) = Q_m(s_t, a_t=A)$.

\subsection{Preliminary Evaluation \& Results}

To evaluate the RMAB model, we compute the Whittle Index for each beneficiary in the test set on the first day of interventions. This is done by using the parameters of the cluster that the beneficiary belongs to, and computing the state $s_t$ using the previous one month of call logs for the beneficiary. We then use the computed action value functions to calculate the Whittle Index for each beneficiary.

To evaluate the performance of the model, we observe the behavior of the top-$k$ beneficiaries flagged for intervention through the RMAB model. Formally, the top-$k$ beneficiaries obtained by sorting them on their Whittle Indices from highest to lowest should give us the subset of beneficiaries that are likely to exhibit high engagement behavior if an intervention is made and low engagement behavior in the absence of an intervention. Hence we consider the top-k beneficiaries of both the call group and the control group. Given that the beneficiaries in the call group received an intervention, we would expect the top-k beneficiaries to exhibit high engagement behavior in the post-intervention period. And given that the beneficiaries in the control group did not receive an intervention, we would expect the top-k beneficiaries to exhibit low engagement behavior in the post-intervention period. Hence, we evaluate the system on the difference of percentage overlaps of top-k beneficiaries based on calculated Whittle Indices and high engagement groups in the call and control groups. A higher difference corresponds to the model identifying a larger set of beneficiaries that exhibit high engagement behavior only as a result of an intervention.

\begin{table}[]
\caption{Intervention Planning Results}
\label{tab:planning-results}
\begin{tabular}{|c|c|c|c|}
\hline
\textbf{Method\footnotemark} & \textbf{Call} & \textbf{Control} \\ \midrule
{\textbf{RMAB}} & $33.8 \pm 5.6$\% & $19.9 \pm 4.6$\% \\ \hline
\end{tabular}
\end{table}

\footnotetext[2]{This evaluation is only indicative of the types of beneficiaries selected for intervention and does not provide a full policy evaluation. A full policy evaluation would require a new intervention study.}

Table \ref{tab:planning-results} shows our results for the RMAB approach. Here, we calculate the percentage overlap of top 100 beneficiaries based on Whittle Index and high engagement groups in the call and control groups. Following the definitions in Section 5.2, high engagement behavior corresponds to an E2C ratio of greater than 0.5 in the post-intervention period. The numbers reported in the table are averaged over 50 runs.

We observe that 33.8\% out of the top 100 beneficiaries that were flagged for a call intervention by our RMAB model exhibited high engagement behavior in the post-intervention period.
The RMAB model is also flagging 19.9\% of beneficiaries in top 100 who have high post-intervention behaviour in the control group. Since beneficiaries in the control group did not receive any intervention, the RMAB model is selecting a lower percentage of those beneficiaries that become high engagement on their own. So overall, our model is useful in identifying beneficiaries that exhibit high engagement behavior only as a result of the intervention.

\section{Discussion}

We have successfully built machine learning models to aid health care workers in NGOs such as ARMMAN to identify women at risk of dropping out of call-based health awareness service. Leveraging call history data collected from the past we have built models that provide high recall for the long-term engagement prediction task. We present the results of the deployment of this model in the field and discuss the effects of the subsequent interventions. We then use data generated from the interventions to develop a method to identify a smaller subset of beneficiaries that are most likely to benefit from an intervention.
As part of our future work, we aim to find the important set of features that lead to the predictions of our model using interpretation techniques like \cite{selvaraju2017grad, merrick2019explanation}. 

\section{Acknowledgements}

We thank Jahnvi Patel and Yogesh Tripathi for their assistance with analysing the data. We also thank Komal Gholekar, Sonali Nandlaskar, Janhvi Dange and Krutika Gosavi for their support with conducting the call interventions. This research was supported by Google AI for Social Good faculty award to Balaraman Ravindran, (Ref: RB1920CP928GOOGRBCHOC).

\bibliographystyle{ACM-Reference-Format} 
\bibliography{sample}


\end{document}